\documentclass{article}

\usepackage[main,final]{neurips_2026}
\makeatletter
\renewcommand{\@noticestring}{}
\makeatother
\usepackage[utf8]{inputenc} % allow utf-8 input
\usepackage[T1]{fontenc}    % use 8-bit T1 fonts
\usepackage{hyperref}       % hyperlinks
\usepackage{url}            % simple URL typesetting
\usepackage{booktabs}       % professional-quality tables
\usepackage{amsfonts}       % blackboard math symbols
\usepackage{nicefrac}       % compact symbols for 1/2, etc.
\usepackage{microtype}      % microtypography
\usepackage{xcolor}         % colors
\usepackage{graphicx}
\usepackage{amsmath,amssymb}

\title{Anticipate Before Acting: Future-State-Conditioned Vision-Language Navigation}

\author{%
  Lingfeng Zhang \thanks{Corresponding to lingfeng.zhang1@h-partners.com, zhanguang.zhang@huawei.com, and yingxue.zhang@huawei.com}\\
  Noah's Ark Lab, 2012 Labs, Huawei \\
  \And
  Zhanguang Zhang \\
  Noah's Ark Lab, 2012 Labs, Huawei \\
  \And
  Liheng Ma \\
  Noah's Ark Lab, 2012 Labs, Huawei \\
  \And
  Tongtong Cao \\
  Department of Foundation model, 2012 Labs, Huawei \\
  \And
  Yingxue Zhang \\
  Noah's Ark Lab, 2012 Labs, Huawei \\
}

\begin{document}

\maketitle

\begin{abstract}
End-to-end vision-language navigation (VLN) with causal vision-language models maps instructions and egocentric observations directly to actions, but standard behavior cloning supervises only the next action and does not explicitly encourage the policy state to be predictive of future visual outcomes, limiting long-horizon decision making.
A \emph{privileged-input diagnostic} shows that access to an \emph{expert-trajectory future image} can substantially improve navigation, indicating that future observations contain rich, actionable cues, though such inputs are unavailable at deployment.
Motivated by this signal, we propose \textbf{Future-State-Conditioned VLN (FSC-VLN)}, a deployable model that augments a causal policy with a \emph{future-query token} and uses \emph{training-only future-state supervision} to distill information from future observations into the policy state.
Concretely, during training we align the future-query representation to a frozen visual embedding $\Delta$ steps ahead, while inference requires only past and current observations. This design preserves the baseline inference pattern and adds only two learned prefix tokens, implying minimal overhead.
On R2R val-unseen, FSC-VLN improves SR/OSR/SPL over a StreamVLN-style baseline under two training-data regimes, with larger gains on long-horizon episodes; ablations further support the dual-query design that separates future and action queries.
\end{abstract}

\section{Introduction}
\label{sec:introduction}

Vision-language navigation (VLN) requires an embodied agent to follow a natural-language instruction by grounding it in a sequence of egocentric visual observations~\citep{anderson2018vision,ku2020room}. In continuous environments (VLN-CE), the agent must further translate this grounding into low-level actions without relying on a predefined navigation graph~\citep{krantz2020beyond}. The resulting policy must jointly maintain instruction progress, interpret a continuously changing scene, and recover from errors accumulated over a long sequence of primitive actions.

Recent vision-language models have enabled a more direct approach
to this problem. Methods such as NaVid~\citep{zhang2024navid}, Uni-NaVid~\citep{zhang2024uni}, NaVILA~\citep{cheng2024navila}, and StreamVLN~\citep{wei2025streamvln} encode the instruction, current observation, and visual history into a multimodal token sequence and fine-tune a pretrained language model to generate navigation actions autoregressively. This formulation reduces reliance on task-specific maps and waypoint-selection modules, but its performance remains strongly tied to large-scale behavior-cloning data. Adding more demonstrations can improve action imitation~\citep{zhang2024uni}, yet it does not identify which missing learning signal limits further progress.

In particular, the standard behavior-cloning objective directly supervises the expert action sequence but does not explicitly require the internal policy state to represent what the agent is likely to observe in the future. A causal VLM may learn such information implicitly from demonstrations, but next-action supervision alone provides no direct constraint on the predictive content of its hidden states. This distinction may matter in continuous navigation: locally plausible actions can still lead to gradual trajectory deviation, and small errors become increasingly difficult to correct over longer execution
horizons. We therefore ask:

\begin{quote}
\emph{Can future-derived training supervision improve a deployable online policy without accessing future observations during inference?}
\end{quote}

We separate the two questions:

\textbf{Diagnostic question:} if a policy is directly given the expert-trajectory future image $\Delta=32$ steps ahead as \emph{privileged input} at test time (and trained with the same privilege), does that additional visual evidence help to choose the current action? This setting is not deployable; we use it only as a sanity check that future observations contain actionable cues for navigation (Sec.~\ref{sec:experiments}, Table~\ref{tab:overall-oracle-headroom}).
% Importantly, the provided ``future'' image is defined along the \emph{expert} trajectory: it is well-defined for the expert rollout, but it is generally not the unique future that an online policy will experience, since small action deviations can lead to different states and observations.

\textbf{Deployable question:} without accessing future images at inference, can we still improve an online policy by using a \emph{compressed future visual latent} as training-only supervision? Because the expert-future observation reflects one particular trajectory outcome and the true on-policy future is action-dependent and uncertain at decision time, this privileged-input diagnostic is meant to establish usefulness rather than to define a target that a deployable method should match. Instead, it motivates learning a representation that is predictive of future visual states while keeping inference unchanged.

Based on this observation, we introduce Future-State-Conditioned VLN (\textbf{FSC-VLN}), a lightweight extension to an end-to-end causal VLM navigation policy. For every observation turn, FSC-VLN inserts one learnable future query followed by one learnable action query into the causal
token sequence. During training, the contextualized hidden state at the future query is aligned with a pooled visual representation of the observation $\Delta$ steps ahead. The target representation is extracted by the shared frozen visual encoder and is treated as a stop-gradient target. Placing the future query before the action query allows
the action-query representation and subsequent action tokens to attend to the future-supervised query prefix through causal self-attention.

The proposed mechanism operates at the token level and does not alter the Transformer blocks, visual encoder architecture, or full-vocabulary language modeling head. The future-target branch is required only during training. At inference, the multimodal context and both learned queries are processed jointly in a single causal prefill. The future query is not decoded autoregressively, nor is its final representation extracted and re-inserted through a separate model invocation. Only the navigation actions are decoded autoregressively. Consequently, FSC-VLN requires no future observation, future-image generator, auxiliary target encoder, or second policy forward pass at inference, and adds only two learned prefix tokens to the original online sequence.

Our experiments answer both questions: Table~\ref{tab:overall-oracle-headroom} addresses the privileged-input diagnostic, while Tables~\ref{tab:main_results}--\ref{tab:stratified-short-long} evaluate the deployable future-latent supervision and show larger gains on long-horizon episodes.

Our contributions are threefold:

\begin{itemize}

    \item We introduce a \emph{privileged expert-future input diagnostic} for
    end-to-end VLM-based VLN. By feeding the expert-trajectory observation $\Delta$
    steps ahead as an additional input during training and evaluation, we
    probe whether future visual evidence can inform the current navigation
    decision. This diagnostic is not deployable and is reported only to
    motivate training-time supervision without future inputs at test time.

    \item We formulate future-state prediction as an auxiliary training
    objective for end-to-end VLM navigation by aligning a future-query
    representation with the frozen visual embedding of an observation $\Delta$
    steps ahead, and realize it via a lightweight dual-query design in which a
    future-supervised query precedes an action query in the causal token
    sequence, enabling future-conditioned action generation without modifying
    the pretrained backbone architecture.

    \item We evaluate FSC-VLN under two training-data regimes and find that its gains are concentrated on long-horizon trajectories, and ablations support the need to separate the future-supervised query from the action-decoding query. At inference, the online context and both queries are processed jointly in a single causal prefill, requiring no autoregressive future-latent generation and no additional policy forward pass. Since the method adds only two prefix tokens per step, the inference cost is expected to remain close to the baseline.

\end{itemize}

\section{Related Work}
\label{sec:related_work}

\paragraph{Vision-language models for navigation.}
Recent vision-language navigation methods increasingly formulate navigation
as action prediction using pretrained multimodal models.
NaVid conditions a video-based VLM on an instruction and an online monocular
RGB stream to predict the next navigation action~\citep{zhang2024navid}.
Uni-NaVid extends this formulation across multiple embodied navigation tasks
by harmonizing their input and low-level action spaces~\citep{zhang2024uni}.
NaVILA instead adopts a hierarchical design in which a VLA model produces
spatially grounded mid-level commands that are executed by a learned
locomotion controller~\citep{cheng2024navila}.
MapNav replaces raw observation history with an annotated top-down semantic
map before predicting navigation actions with a VLM~\citep{zhang2025mapnav}.
More recent methods focus on representing long observation histories
efficiently.
StreamVLN maintains a slow-updating visual memory together with a
fast-streaming dialogue context~\citep{wei2025streamvln}, whereas JanusVLN
separates visual-semantic and spatial-geometric information into two compact
implicit memories~\citep{zeng2025janusvln}.

Although these methods differ in their memory representations and action
interfaces, their policy learning remains predominantly action-centric:
the multimodal context is optimized to predict an action or an intermediate
action command, without explicitly training a policy state to match a future
visual observation and subsequently using that state to condition action
prediction. Our work studies whether adding such predictive structure to an
end-to-end VLM navigator improves decision making, particularly over long
navigation horizons.

\paragraph{Future prediction for embodied control.}
Predictive visual modeling has recently shown substantial promise in robotic
manipulation.
DreamZero jointly models future video and robot actions using an
autoregressive video-diffusion backbone, demonstrating that video dynamics
pretraining can improve cross-task and cross-embodiment generalization~\citep{ye2026world}.
LingBot-VA similarly interleaves video and action representations in an
autoregressive diffusion model and jointly predicts visual evolution and
control trajectories~\citep{li2026causal}.
These approaches make future imagination an explicit part of the
video-action model, but iterative video denoising can introduce considerable
deployment costs.

FastWAM separates training-time world modeling from test-time imagination:
it retains future-video co-training but directly predicts actions at
inference without generating future observations~\citep{yuan2026fast}.
Its controlled experiments suggest that much of the control benefit comes
from the representation learned through predictive supervision, rather than
from explicitly rendering a future rollout at test time.
VLA-JEPA further shifts \textit{future prediction} from pixel space to representation space: it encodes future video frames into latent targets with a frozen encoder and trains a predictor (student pathway) to match these future latents given only past and current context. Importantly, the raw future frames are used only to form the stop-gradient targets and are not provided as inputs to the policy at test time~\citep{sun2026vla}. H-WM jointly models high-level logical transitions and low-level latent visual subgoals, providing hierarchical symbolic–perceptual guidance for long-horizon robotic manipulation~\citep{huang2026h}. LingBot-VLA 2.0 further introduces current and future queries that are
distilled from complementary depth and causal video teachers~\citep{wu2026foundation}.
Together, these results provide strong evidence that future-image or
future-latent supervision can improve manipulation policies.
However, they do not directly establish whether the same principle benefits
instruction-guided navigation, where errors accumulate over substantially
longer action sequences, and the agent must maintain consistency between
language, partial observations, and spatial progress.

\section{Method}
\label{sec:method}

\subsection{Problem Formulation}
\label{sec:problem_formulation}

At navigation step $t$, the agent receives a natural-language instruction
$\mathcal{I}$, a current visual observation $o_t$, and a sequence of
historical observations $\mathcal{H}_t$. The policy predicts a short sequence
of navigation actions
\begin{equation}
    \mathbf{a}_t =
    \left(a_{t,1}, \ldots, a_{t,M}\right),
\end{equation}
and each action belongs to
\begin{equation}
    \mathcal{A}
    =
    \left\{
        % \texttt{MoveForward},
        % \texttt{TurnLeft},
        % \texttt{TurnRight},
        \uparrow,
        \leftarrow,
        \rightarrow,
        \texttt{STOP}
    \right\}.
\end{equation}
Actions are represented as ordinary language tokens and generated
autoregressively by a causal language model. Following StreamVLN, navigation actions are represented as text tokens and
decoded autoregressively through the full-vocabulary language model (LM) head.

During training, we additionally observe a future image $o_{t+\Delta}$. The future image is used only to supervise an
internal future-state representation. It is never provided to the online policy and is not required at inference time.

\subsection{Model Overview}
\label{sec:model_overview}

\begin{figure}
  \centering
  \includegraphics[width=1.0\linewidth]{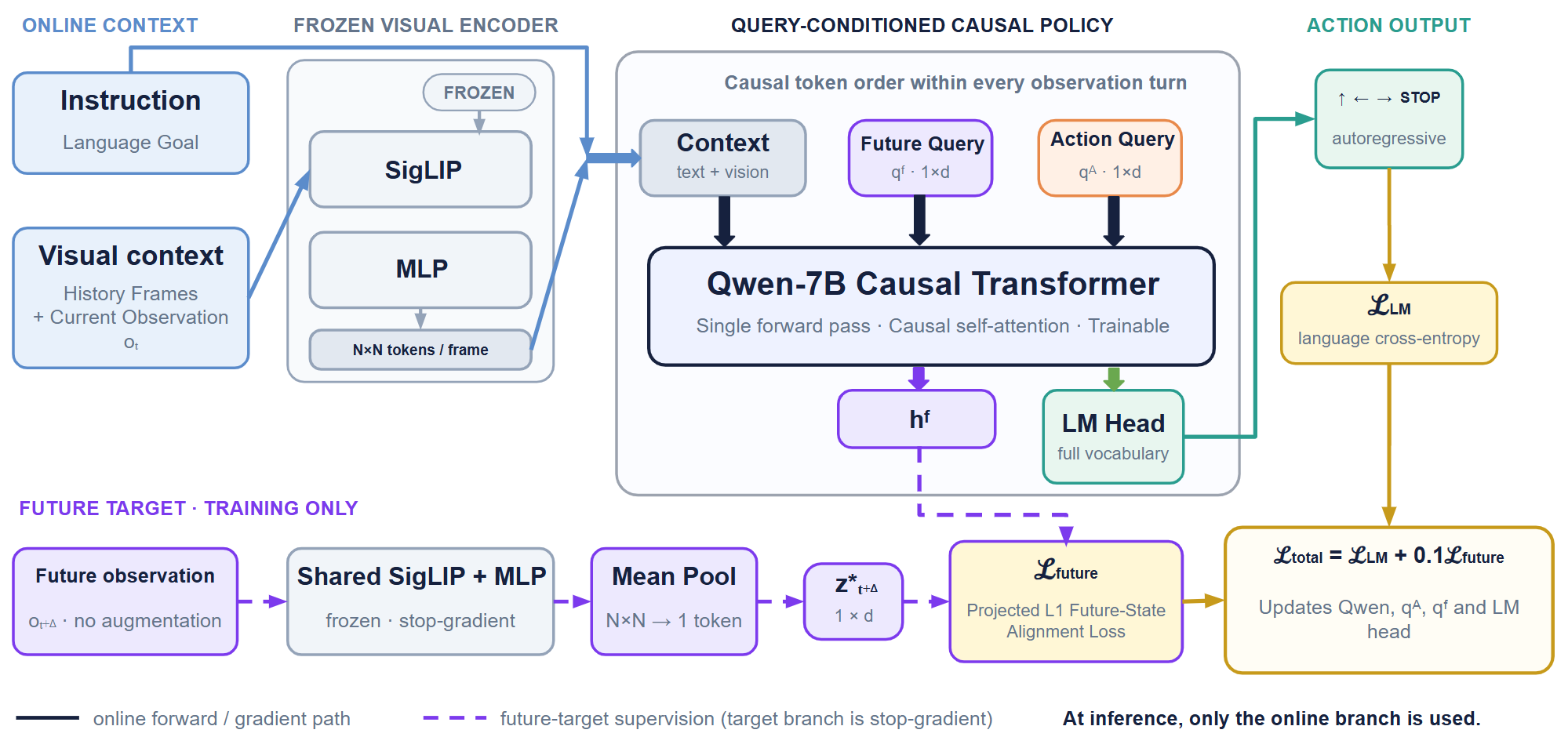}
  \caption{Overview of FSC-VLN. A frozen visual encoder maps the current and historical observations into visual tokens that, together with the instruction, form the online context. We prepend a learnable future query $\mathbf{q}^{F}$ and an action query $\mathbf{q}^{A}$ to create a future-supervised causal prefix for action generation. During training only, a target branch encodes the expert-trajectory future observation $o_{t+\Delta}$ with the same frozen visual modules to provide a stop-gradient embedding target for aligning $\mathbf{q}^{F}$; this branch is discarded at inference.}
  \label{fig:method_overview}
\end{figure}

Figure~\ref{fig:method_overview} presents the architecture of our
future-state-conditioned policy. The model contains three components: a frozen visual encoder, a query-conditioned causal policy, and a training-only future-target branch.

The online branch encodes the instruction, historical observations, and current observation into a multimodal context. Two learnable query tokens are inserted before action generation: a future query $\mathbf{q}^{F}$, which
predicts a latent representation of a future observation, followed by an action query $\mathbf{q}^{A}$. Action generation is conditioned on a future-supervised causal prefix. Both queries are processed jointly with the multimodal context by a Qwen2-7B~\citep{yang2024qwen2} causal Transformer.

The future-target branch encodes $o_{t+\Delta}$ using the same frozen visual modules as the online branch. Its output provides a stop-gradient target for the future query. The policy therefore learns to anticipate a future visual state while retaining a single causal Transformer pass for online action
prediction.

\subsection{Frozen Visual Encoding}
\label{sec:visual_encoding}

Following StreamVLN~\citep{wei2025streamvln}, we use SigLIP~\citep{zhai2023sigmoid} as the visual encoder. Given
an image $o$, the encoder produces a $M \times M$ grid of patch features. A
two-layer MLP with GELU activation maps each feature from the SigLIP feature
dimension of $d^*$ to the language-model dimension $d$:
\begin{equation}
    \widetilde{\mathbf{V}}(o)
    =
    P_{\mathrm{vis}}
    \left(
        E_{\mathrm{vis}}(o)
    \right)
    \in
    \mathbb{R}^{M^2 \times d},
    \label{eq:visual_projection}
\end{equation}
where $E_{\mathrm{vis}}$ denotes SigLIP and $P_{\mathrm{vis}}$ denotes the projector.

We apply mean pooling with a stride of two, reducing the feature grid from
$M \times M$ to $N \times N$. Each frame is therefore represented by
$N^2$ visual tokens:
\begin{equation}
    \mathbf{V}(o)
    \in
    \mathbb{R}^{N^2 \times d}.
    \label{eq:visual_tokens}
\end{equation}
The visual tokens from the current observation and historical frames are
interleaved with the instruction tokens to construct the online context
$\mathbf{C}_t$. Both $E_{\mathrm{vis}}$ and $P_{\mathrm{vis}}$ remain frozen
throughout training. This keeps the visual target space stationary and
concentrates optimization on the causal policy and query representations.

\subsection{Query-Conditioned Causal Policy}
\label{sec:query_conditioned_policy}

For each observation turn, we append one learnable future query and one learnable action query to
the multimodal context. Omitting fixed dialogue delimiter tokens for clarity, the resulting causal sequence is
\begin{equation}
    \mathbf{X}_t
    =
    \left[
        \mathbf{C}_t,\,
        \mathbf{q}^{F},\,
        \mathbf{q}^{A},\,
        \mathbf{a}_t
    \right],
    \label{eq:causal_sequence}
\end{equation}
where
\begin{equation}
    \mathbf{q}^{F},
    \mathbf{q}^{A}
    \in
    \mathbb{R}^{1 \times d}.
\end{equation}
Both queries are trainable parameters initialized from $\mathcal{N}(\mu,\sigma^2)$.

% $\mathcal{N}(0,0.02^2)$.

The complete sequence is processed by the causal Transformer $F_{\phi}$:
\begin{equation}
    \mathbf{H}_t
    =
    F_{\phi}\left(\mathbf{X}_t\right).
    \label{eq:causal_transformer}
\end{equation}
We denote the hidden states at the future-query and action-query positions by
$\mathbf{h}_t^{F}$ and $\mathbf{h}_t^{A}$, respectively. Under causal
self-attention, $\mathbf{h}_t^{F}$ summarizes the instruction and visual
context:
\begin{equation}
    \mathbf{h}_t^{F}
    =
    F_{\phi}
    \left(
        \mathbf{C}_t,
        \mathbf{q}^{F}
    \right).
    \label{eq:future_query_hidden}
\end{equation}
The action query can additionally attend to the future-query representation:
\begin{equation}
    \mathbf{h}_t^{A}
    =
    F_{\phi}
    \left(
        \mathbf{C}_t,
        \mathbf{q}^{F},
        \mathbf{q}^{A}
    \right)
    \label{eq:action_query_hidden}
\end{equation}
This ordering provides an explicit causal path from future-state prediction
to action generation. The action query does not receive an expert-trajectory future
feature; it only accesses the context-dependent representation computed at the
future-query position.

The Transformer hidden states are mapped to full-vocabulary logits by the
language-model head:
\begin{equation}
    \boldsymbol{\ell}_i
    =
    \mathbf{W}_{\mathrm{LM}}\mathbf{h}_i,
    \label{eq:lm_head}
\end{equation}
where $\mathbf{h}_i$ denotes the Transformer hidden state at position $i$ and
\begin{equation}
    \mathbf{W}_{\mathrm{LM}}
    \in
    \mathbb{R}^{|\mathcal{V}| \times d}
\end{equation}

In particular, omitting fixed dialogue-template tokens, the logits for the
first action are obtained from the action-query hidden state:
\begin{equation}
    \boldsymbol{\ell}_{t,1}
    =
    \mathbf{W}_{\mathrm{LM}}\mathbf{h}_t^{A}
    \label{eq:first_action_logits}
\end{equation}
The subsequent action tokens are predicted autoregressively:
\begin{equation}
    \boldsymbol{\ell}_{t,j}
    =
    \mathbf{W}_{\mathrm{LM}}
    \mathbf{h}_{t,j-1}^{\mathrm{act}}, \qquad j>1
    \label{eq:subsequent_action_logits}
\end{equation}
where $\mathbf{h}_{t,j-1}^{\mathrm{act}}$ is the hidden state associated
with the preceding generated action token.

The LM head is a bias-free linear layer. The probability of the next token is
\begin{equation}
\begin{split}
    p_{\theta}
    \left(
        y_{i+1}
        \mid
        y_{\leq i},
        \mathbf{C}_t,
        \mathbf{q}^{F},
        \mathbf{q}^{A}
    \right) =
    \operatorname{softmax}
    \left(
        \mathbf{W}_{\mathrm{LM}}\mathbf{h}_i
    \right)
\end{split}
\label{eq:next_token_probability}
\end{equation}
The policy generates the textual action symbols
$\texttt{STOP}$, $\uparrow$, $\leftarrow$, and $\rightarrow$
autoregressively. At execution time, the generated text is parsed into the
corresponding discrete navigation actions.

\subsection{Future-State Target Construction}
\label{sec:future_target}

The future target is constructed from the observation $o_{t+\Delta}$. During training, stochastic appearance augmentation is applied only to the online context images. Future target images undergo the standard SigLIP preprocessing without additional stochastic augmentation, providing a stable
target for future-state alignment. It is passed through the shared frozen visual encoder and projector:
\begin{equation}
    \mathbf{V}_{t+\Delta}^{*}
    =
    \operatorname{sg}
    \left[
        \mathbf{V}(o_{t+\Delta})
    \right]
    \in
    \mathbb{R}^{N^2 \times d},
    \label{eq:future_visual_target}
\end{equation}
where $\operatorname{sg}[\cdot]$ denotes the stop-gradient operation.

Because the policy uses a single future query, we aggregate the $N^2$ future
visual tokens using mean pooling:
\begin{equation}
    \mathbf{z}_{t+\Delta}^{*}
    =
    \frac{1}{N^2}
    \sum_{j=1}^{N^2}
    \mathbf{V}_{t+\Delta,j}^{*}
    \in
    \mathbb{R}^{d}
    \label{eq:future_target_pooling}
\end{equation}
The target branch shares the frozen SigLIP and MLP parameters with the online
branch. Consequently,
$\mathbf{z}_{t+\Delta}^{*}$ defines a deterministic target in the embedding space.

The predicted future state is the Transformer output at the future-query
position:
\begin{equation}
    \widehat{\mathbf{z}}_{t+\Delta}
    =
    \mathbf{h}_t^{F}
    \label{eq:predicted_future_state}
\end{equation}

\subsection{Projected Future-State Alignment}
\label{sec:future_alignment}

The future-state target $\mathbf{z}_{t+\Delta}^{*}\in\mathbb{R}^{d}$ is high-dimensional, so we perform the alignment in a compact projection space. Let $g:\mathbb{R}^{d}\rightarrow\mathbb{R}^{d'}$ denote a projection with $d'\ll d$. We define the projected future-state loss as
\begin{equation}
    \mathcal{L}_{\mathrm{future}}
    =
    \left\lVert
        g\left(\widehat{\mathbf{z}}_{t+\Delta}\right)
        -
        g\left(\mathbf{z}_{t+\Delta}^{*}\right)
    \right\rVert_{1}
    \label{eq:future_loss}
\end{equation}
We use an $\ell_1$ penalty to reduce sensitivity to individual feature outliers in the projected space.

Because the target pathway applies stop-gradient, gradients do not flow into the frozen visual encoder:
\begin{equation}
    \nabla_{\mathbf{z}_{t+\Delta}^{*}}\,\mathcal{L}_{\mathrm{future}}=\mathbf{0}
\end{equation}
As a result, $\mathcal{L}_{\mathrm{future}}$ updates the causal Transformer parameters and the future-query embedding, while leaving the target encoder unchanged. Gradients from action prediction can additionally reach the future query through the causal dependency $\mathbf{q}^{F}\rightarrow\mathbf{q}^{A}\rightarrow\mathbf{a}_t$, encouraging the future-query representation to be both predictive of the expert future observation and useful for action generation.

\subsection{Training Objective}
\label{sec:training_objective}

The navigation policy is trained using standard next-token cross-entropy over
the full language-model vocabulary. Let $\mathbf{Y}$ denote the tokenized
training sequence and let $\mathcal{S}$ be the set of supervised
assistant-side positions. The language-modeling loss is
\begin{equation}
    \mathcal{L}_{\mathrm{LM}}
    =
    -
    \frac{1}{|\mathcal{S}|}
    \sum_{i\in\mathcal{S}}
    \log
    p_{\theta}
    \left(
        Y_i
        \mid
        Y_{<i},
        \mathbf{C}_t,
        \mathbf{q}^{F},
        \mathbf{q}^{A}
    \right)
    \label{eq:lm_loss}
\end{equation}

The complete training objective is
\begin{equation}
    \mathcal{L}_{\mathrm{total}}
    =
    \mathcal{L}_{\mathrm{LM}}
    +
    \lambda_{F}
    \mathcal{L}_{\mathrm{future}}
    \label{eq:total_loss}
\end{equation}
We optimize the causal language model (including the LM head) together with the future and action query embeddings. The SigLIP encoder and the MLP projector remain frozen.

\subsection{Inference}
\label{sec:inference}

At inference time, the future-target branch is discarded. The multimodal
context, future query, and action query are concatenated and processed
jointly in a single causal prefill. The future query is not decoded
autoregressively; its contextualized hidden state serves as the predicted
future-state representation. Since the future query precedes the action query
in the causal sequence, the action-query representation and subsequent action
tokens can attend to the future-query prefix. The LM head then decodes the
navigation actions autoregressively. No future observation, target feature, or auxiliary target encoder is required
at inference time.

\section{Experiments}
\label{sec:experiments}

\paragraph{Setup}
We use StreamVLN as our baseline end-to-end VLM navigation policy. All shared components (backbone, optimizer, learning-rate schedule, batch size, data preprocessing, and decoding settings) follow the StreamVLN configuration to ensure a fair comparison.

\paragraph{Metrics}
We report Success Rate (SR), the fraction of episodes in which the agent stops within a goal tolerance; Oracle Success Rate (OSR), the fraction of episodes whose trajectory ever comes within the goal tolerance (regardless of the final stopping point); and Success weighted by Path Length (SPL), which complements success by accounting for navigation efficiency via the ratio between the shortest-path distance and the executed path length.

\paragraph{Data}
We train and evaluate on standard indoor VLN benchmarks: \textbf{R2R} (Room-to-Room)~\citep{krantz2020beyond} and \textbf{RxR} (Room-across-Room)~\citep{ku2020room}, which provide human-annotated language instructions paired with navigation trajectories in photorealistic indoor environments. Unless otherwise stated, results are reported on the R2R \texttt{val-unseen} split. Following prior practice in StreamVLN-style training, we also consider an expanded training regime that augments R2R+RxR with \textbf{EnvDrop}~\citep{tan2019learning}, a data augmentation that increases environment variability during training, and \textbf{ScaleVLN}~\citep{wang2023scaling}, an additional large-scale VLN training set used to broaden demonstration coverage.

\paragraph{Trajectory Stratification}
To analyze horizon effects, we stratify episodes by the geodesic distance between the start and goal locations: episodes with geodesic distance $\leq 10\,\mathrm{m}$ are labeled \emph{short-horizon}, and those with distance $> 10\,\mathrm{m}$ are labeled \emph{long-horizon}. On R2R \texttt{val-unseen} (1839 episodes), this yields 1293 short-horizon and 546 long-horizon episodes.

\paragraph{Implementation details}
We initialize the policy from \texttt{Qwen2} and use \texttt{SigLIP-SO400M} as the frozen visual encoder. Each training sample is a trajectory segment of up to 32 low-level control steps; the policy predicts four actions per observation turn. We use one learnable future query and one learnable action query.

For future supervision, the target observation is chosen at offset $\Delta=32$ steps along the expert trajectory from the current step; if $t+\Delta$ exceeds the trajectory length, we use the final frame. Stochastic appearance augmentation is applied only to the online context images, while the future target uses standard SigLIP preprocessing.

We supervise the future-query hidden state with the projected $\ell_1$ alignment loss (Sec.~\ref{sec:future_alignment}) and set $\lambda_{F}=0.1$. We train for one epoch over the constructed trajectory segments. Training uses 4 GPUs with per-device batch size 1 and gradient accumulation of 16 steps (global batch size 64). We use AdamW with learning rate $2\times 10^{-5}$, weight decay 0, warmup ratio 0.075, a cosine schedule, and gradient clipping at 1.0.

We fine-tune the \texttt{Qwen2} causal language model and LM head together with the future/action query embeddings, while keeping the SigLIP encoder and visual MLP projector frozen. Training uses bfloat16 precision, gradient checkpointing, and DeepSpeed ZeRO-2. We use the same optimization settings for both the R2R+RxR and R2R+RxR+EnvDrop+ScaleVLN training configurations.

\paragraph{Privileged expert-future input diagnostic}
We first answer the \emph{diagnostic question}: if the policy is given the expert-trajectory future observation $\Delta=32$ steps ahead along the expert trajectory as an additional input \emph{at test time} (and trained with the same privilege), is that future visual evidence useful for choosing the current action? This setting is not deployable and is used only as a sanity check. The result shows that future observations can be highly informative, improving SR from $29.96\%$ to $71.77\%$, OSR from $38.49\%$ to $74.00\%$, and SPL from $26.25\%$ to $64.15\%$ (Table~\ref{tab:overall-oracle-headroom}).

\begin{table}
  \caption{Privileged expert-future input diagnostic on R2R \texttt{val-unseen} (1839 episodes). The future image is provided as an additional policy input during both training and evaluation. This experiment tests whether expert-future visual information is useful for navigation and is not an upper bound on FSC-VLN. We train on (R2R+RxR). $\Delta$ denotes absolute change (percentage points).}
  \label{tab:overall-oracle-headroom}
  \centering
  \begin{tabular}{llll}
    \toprule
    Model & SR & Oracle SR & SPL \\
    \midrule
    Baseline (R2R + RxR) & 29.96\% & 38.49\% & 26.25\% \\
    Baseline + Expert-future image at train and test (diagnostic) & 71.77\% & 74.00\% & 64.15\% \\
    Gain $\Delta$ & +41.81 & +35.51 & +37.90 \\
    \bottomrule
  \end{tabular}
\end{table}

\paragraph{Main Results}

As shown in Table~\ref{tab:main_results}, across both training-data regimes in our experiments, FSC-VLN improves all three aggregate navigation metrics. With R2R+RxR training data, it improves SR from $29.96\%$ to $31.43\%$, OSR from $38.49\%$ to $41.76\%$, and SPL from $26.25\%$ to $27.36\%$, corresponding to gains of $1.47$, $3.27$, and $1.11$ percentage points. When the training set is expanded with EnvDrop and ScaleVLN, FSC-VLN improves SR, OSR, and SPL by $0.98$, $1.04$, and $0.85$ percentage points, respectively.

\begin{table}
  \caption{Overall performance. We train on (R2R+RxR+\dots) and evaluate on R2R \texttt{val-unseen} (1839 episodes). $\Delta$ denotes absolute change (percentage points).}
  \label{tab:main_results}
  \centering
  \begin{tabular}{llll}
    \toprule
    Model & SR & Oracle SR & SPL \\
    \midrule
    Baseline (R2R + RxR) & 29.96\% & 38.49\% & 26.25\% \\
    FSC-VLN (ours) & 31.43\% & 41.76\% & 27.36\% \\
    Gain $\Delta$ & +1.47 & +3.27 & +1.11 \\
    \midrule
    Baseline (R2R + RxR + EnvDrop + ScaleVLN) & 44.81\% & 49.97\% & 41.96\% \\
    FSC-VLN (ours) & 45.79\% & 51.01\% & 42.81\% \\
    Gain $\Delta$ & +0.98 & +1.04 & +0.85 \\
    \bottomrule
  \end{tabular}
\end{table}

As shown in Table~\ref{tab:stratified-short-long}, the aggregate results conceal a pronounced dependence on trajectory length. Under R2R+RxR training, FSC-VLN changes short-horizon SR and SPL by only $+0.46$ and $+0.02$ points, while improving long-horizon SR, OSR, and SPL by $3.85$, $7.15$, and $3.70$ points. The same pattern remains after adding EnvDrop and ScaleVLN: short-horizon SR and SPL change by $+0.15$ points each, whereas long-horizon SR, OSR, and SPL improve by $2.93$, $3.67$, and $2.52$ points. The method therefore leaves short-horizon behavior largely unchanged, while its observed gains are concentrated on longer trajectories. This stratified result supports the hypothesis that future-state supervision is most useful when local decisions must remain coherent over an extended
execution horizon.

\begin{table}
  \caption{Stratified short- vs long-horizon evaluation. We train on (R2R+RxR+\dots) and evaluate on R2R \texttt{val-unseen} (1293 short / 546 long). $\Delta$ denotes absolute change (percentage points).}
  \label{tab:stratified-short-long}
  \centering
  \small
  \setlength{\tabcolsep}{3pt}
  \begin{tabular}{p{5cm}cccccc}
    \toprule
    & \multicolumn{3}{c}{Short-horizon} & \multicolumn{3}{c}{Long-horizon} \\
    Model & SR & Oracle SR & SPL & SR & Oracle SR & SPL \\
    \midrule
    \shortstack[l]{Baseline (R2R + RxR)} & 33.49\% & 43.39\% & 29.29\% & 21.61\% & 26.92\% & 19.07\% \\
    \shortstack[l]{FSC-VLN (ours)} & 33.95\% & 45.01\% & 29.31\% & 25.46\% & 34.07\% & 22.77\% \\
    Gain $\Delta$ & +0.46 & +1.62 & +0.02 & +3.85 & +7.15 & +3.70 \\
    \midrule
    \shortstack[l]{Baseline (R2R + RxR + EnvDrop + \\ScaleVLN)} & 50.66\% & 57.23\% & 47.22\% & 30.95\% & 32.78\% & 29.51\% \\
    \shortstack[l]{FSC-VLN (ours)} & 50.81\% & 57.15\% & 47.37\% & 33.88\% & 36.45\% & 32.03\% \\
    Gain $\Delta$ & +0.15 & -0.08 & +0.15 & +2.93 & +3.67 & +2.52 \\
    \bottomrule
  \end{tabular}
\end{table}

\paragraph{Ablation Study: Is a separate action query necessary?}
Our model uses two query positions, $q^{F}$ and $q^{A}$, inserted between the
context and the action tokens:
$\mathcal{C}_t \rightarrow q^{F} \rightarrow q^{A} \rightarrow \mathbf{Y}_t$.
We supervise the hidden state at $q^{F}$ with the future-state alignment loss,
while $q^{A}$ attends to $q^{F}$ and serves as the immediate conditioning
state for action decoding.

To test whether $q^{A}$ is needed, we remove it and keep all other settings
(backbone, future target, loss weight, and schedule) unchanged, yielding
$\mathcal{C}_t \rightarrow q^{F} \rightarrow \mathbf{Y}_t$.
In this variant, the same state at $q^{F}$ must both match the future visual
target and provide the prefix state used to predict the first action token.

Table~\ref{tab:stratified-short-long-ablation} shows that removing $q^{A}$ reduces SR (45.79\% $\rightarrow$ 45.19\%) and SPL (42.81\% $\rightarrow$ 41.76\%). The stratified results indicate that the drop is larger on long-horizon episodes (SR: 33.88\% $\rightarrow$ 32.60\%, SPL: 32.03\% $\rightarrow$ 30.94\%).

Overall, the action query improves final path efficiency even when the agent
can occasionally reach the goal region (as reflected by Oracle SR), suggesting
that separating future-state prediction ($q^{F}$) from action conditioning
($q^{A}$) better preserves and uses the predicted future information during
decoding.

% \begin{table}
%   \caption{Ablation study. We train on (R2R+RxR+EnvDrop+ScaleVLN) and evaluate on R2R \texttt{val-unseen} (1839 episodes). $\Delta$ denotes absolute change (percentage points).}
%   \label{tab:main_ablation}
%   \centering
%   \begin{tabular}{llll}
%     \toprule
%     Model & SR & Oracle SR & SPL \\
%     \midrule
%     FSC-VLN (future query + action query) & 45.79\% & 51.01\% & 42.81\% \\
%     FSC-VLN (only future query) & 45.19\% & 51.77\% & 41.76\% \\
%     Diff $\Delta$ & -0.60 & +0.76 & -1.05 \\
%     \bottomrule
%   \end{tabular}
% \end{table}

% \begin{table}
%   \caption{Ablation study on stratified short- vs long-horizon evaluation. We train on (R2R+RxR+EnvDrop+ScaleVLN) and evaluate on R2R \texttt{val-unseen} (1293 short / 546 long). $\Delta$ denotes absolute change (percentage points).}
%   \label{tab:stratified-short-long-ablation}
%   \centering
%   \small
%   \setlength{\tabcolsep}{3pt}
%   \begin{tabular}{p{5cm}cccccc}
%     \toprule
%     & \multicolumn{3}{c}{Short-horizon} & \multicolumn{3}{c}{Long-horizon} \\
%     Model & SR & Oracle SR & SPL & SR & Oracle SR & SPL \\
%     \midrule
%     FSC-VLN (future query + action query) & 50.81\% & 57.15\% & 47.37\% & 33.88\% & 36.45\% & 32.03\% \\
%     FSC-VLN (only future query) & 50.50\% & 58.78\% & 46.33\% & 32.60\% & 35.16\% & 30.94\% \\
%     Diff $\Delta$ & -0.31 & +1.63 & -1.04 & -1.28 & -1.29 & -1.09 \\
%     \bottomrule
%   \end{tabular}
% \end{table}

\begin{table}
  \caption{Ablation on full-episode evaluation, stratified by short vs.\ long horizons. Models are trained on R2R+RxR+EnvDrop+ScaleVLN and evaluated on R2R \texttt{val-unseen} (1839 total: 1293 short, 546 long). $\Delta$ denotes absolute difference in percentage points (pp).}
  \label{tab:stratified-short-long-ablation}
  \centering
  \small
  \setlength{\tabcolsep}{3pt}
  \begin{tabular}{p{5cm}ccccccccc}
    \toprule
    & \multicolumn{2}{c}{Full-episode} & \multicolumn{2}{c}{Short-horizon} & \multicolumn{2}{c}{Long-horizon} \\
    Model & SR & SPL & SR & SPL & SR & SPL \\
    \midrule
    FSC-VLN (future query + action query) & 45.79\% & 42.81\% & 50.81\% & 47.37\% & 33.88\% & 32.03\% \\
    FSC-VLN (only future query) & 45.19\% & 41.76\% & 50.50\% & 46.33\% & 32.60\% & 30.94\% \\
    Diff $\Delta$ & -0.60 & -1.05 & -0.31 & -1.04 & -1.28 & -1.09 \\
    \bottomrule
  \end{tabular}
\end{table}

\section{Discussion}
\label{sec:discussion}

\paragraph{Why future-state supervision helps.}
Behavior cloning optimizes for immediate action prediction, but it does not directly encourage the policy state to be predictive of the downstream visual consequences of earlier decisions. The auxiliary alignment objective adds a direct constraint that ties part of the hidden state to a future observation embedding.

Our ablation results also suggest that \emph{how} this predictive signal is exposed to the decoder matters. When the future-supervised query must simultaneously serve as the action-conditioning state (i.e., removing the separate action query), Oracle SR can increase while SR/SPL decrease, and the drop is more pronounced on long-horizon episodes. This pattern is consistent with a representation-level trade-off: the future-supervised state may carry useful goal-reaching cues, but a dedicated action query helps convert that information into efficient step-by-step decisions through causal attention.

\paragraph{Efficiency and deployment.}
FSC-VLN keeps the baseline inference pattern: one causal prefill over the online context, followed by autoregressive decoding of action tokens. The future-target branch is used only during training, and inference introduces only two learned prefix tokens.

\paragraph{Limitations.}
The future target is taken from an expert trajectory and may differ from the agent's on-policy future under compounding errors. In addition, a fixed horizon $\Delta$ may not be appropriate for all scenes and episode lengths. Finally, our experiments focus on a single backbone; it remains to validate the approach across other VLN backbones and environments.

\paragraph{Future directions.}
Future work could study adaptive or multi-step targets (e.g., multiple $\Delta$ values), uncertainty-aware objectives, and combinations with rollout-level training signals that optimize global instruction completion and path efficiency without changing inference.

\section{Conclusion}
\label{sec:conclusion}

We use a privileged expert-future input diagnostic to verify that future observations contain actionable visual evidence for end-to-end VLM navigation. We then propose FSC-VLN, a deployable method that improves an online policy without accessing future images at inference by aligning a future-query representation to a compressed future visual latent during training. In our experiments on R2R \texttt{val-unseen}, FSC-VLN yields modest overall gains with larger improvements on long-horizon episodes, and ablations indicate that a dedicated action query helps translate future-supervised representations into better navigation behavior.

% NeurIPS template uses natbib; you must specify a .bst file for BibTeX.
\bibliographystyle{plainnat}
\bibliography{reference}

%%%%%%%%%%%%%%%%%%%%%%%%%%%%%%%%%%%%%%%%%%%%%%%%%%%%%%%%%%%%

% \appendix

% \section{Technical appendices and supplementary material}
% Technical appendices with additional results, figures, graphs, and proofs may be submitted with the paper submission before the full submission deadline (see above). You can upload a ZIP file for videos or code, but do not upload a separate PDF file for the appendix. There is no page limit for the technical appendices. 

% Note: Think of the appendix as ``optional reading'' for reviewers. The paper must be able to stand alone without the appendix; for example, adding critical experiments that support the main claims to an appendix is inappropriate. 

% %%%%%%%%%%%%%%%%%%%%%%%%%%%%%%%%%%%%%%%%%%%%%%%%%%%%%%%%%%%%

% \newpage
% \input{checklist.tex}

\end{document}